\documentclass[10pt,twocolumn,letterpaper]{article}

\usepackage{3dv}
\usepackage{times}
\usepackage{epsfig}
\usepackage{graphicx}
\usepackage{amsmath}
\usepackage{amssymb}
\usepackage[table]{xcolor}
\usepackage{booktabs}

\usepackage[pagebackref=true,breaklinks=true,letterpaper=true,colorlinks,bookmarks=false]{hyperref}

\threedvfinalcopy 

\def\threedvPaperID{354} 

\ifthreedvfinal\fi
\newif\ifdraft
\draftfalse
\drafttrue

\definecolor{orange}{rgb}{1,0.5,0}

\ifdraft
 \newcommand{\PF}[1]{{\color{red}{\bf PF: #1}}}
 
 \newcommand{\MS}[1]{{\color{green}{\bf MS: #1}}}
 
 \newcommand{\ZD}[1]{{\color{blue}{\bf ZD: #1}}}
 
 \newcommand{\JB}[1]{{\color{orange}{\bf JB: #1}}}

\else
 \newcommand{\PF}[1]{}
 
 \newcommand{\MS}[1]{}
 
 \newcommand{\ZD}[1]{}
 
 \newcommand{\JB}[1]{}
 
\fi

\newcommand{\comment}[1]{}
\newcommand{\parag}[1]{\vspace{-3mm}\paragraph{#1}}

\newcommand{\real}{\mathbb{R}}

\newcommand{\pb}{\mathbf{p}}
\newcommand{\qb}{\mathbf{q}}
\newcommand{\wb}{\mathbf{w}}
\newcommand{\nb}{\mathbf{n}}

\newcommand{\pbi}{\mathbf{p}_{i}}

\newcommand{\pbhj}{\mathbf{\hat{p}}_{j}}
\newcommand{\pbik}{\mathbf{p}_{k_{i}}}

\newcommand{\fwk}[1]{f_{\mathbf{w}_{#1}}}

\newcommand{\Lchd}{\mathcal{L}_{\text{CHD}}}

\newcommand{\LE}{\mathcal{L}_{E}}
\newcommand{\LG}{\mathcal{L}_{G}}

\newcommand{\Loverlap}{\mathcal{L}_{\text{ol}}}
\newcommand{\Lskew}{\mathcal{L}_{\text{sk}}}

\newcommand{\Eik}{E_{i}^{(k)}}
\newcommand{\Fik}{F_{i}^{(k)}}
\newcommand{\Gik}{G_{i}^{(k)}}
\newcommand{\Ak}{A^{(k)}}

\newcommand{\sumKM}{\frac{1}{KM}\sum_{k=1}^{K}\sum_{i=1}^{M}}
\newcommand{\alphE}{\alpha_{E}}
\newcommand{\alphG}{\alpha_{G}}
\newcommand{\alphskew}{\alpha_{\text{sk}}}

\newcommand{\alpholap}{\alpha_{\text{ol}}}
\newcommand{\Lmain}{\mathcal{L}}

\newcommand{\molap}{m_{\text{olap}}}
\newcommand{\mAE}{m_{\text{ae}}}

\newcommand{\cb}{\mathbf{c}}
\newcommand{\gik}{g_{i}^{(k)}}

\newcommand{\nbkip}{\mathbf{n}_{k_{i}}^{p}}
\newcommand{\nbkig}{\mathbf{n}_{k_{i}}^{g}}

\newcommand{\Lsc}{\mathcal{L}_{\text{sc}}}
\newcommand{\Lst}{\mathcal{L}_{\text{st}}}
\newcommand{\Ldsp}{\mathcal{L}_{\text{DSP}}}
\newcommand{\alphsc}{\alpha_{\text{sc}}}
\newcommand{\alphst}{\alpha_{\text{st}}}
\newcommand{\mS}{m_{S}}

\newcommand{\OURSaprx}{aprox}
\newcommand{\OURSanlt}{analyt}
\newcommand{\OURSanltstch}{analyt+stitch}
\newcommand{\OURSanltarea}{analyt-area}
\newcommand{\OURSanltstcharea}{analyt+stitch-area}
\newcommand{\OURSstitch}{stitch}

\newcommand{\PCAE}{\textbf{PCAE}}

\newcommand{\SVR}{\textbf{SVR}}

\newcommand{\DSP}{DSP}

\newcommand{\OURS}[1]{\text{OURS}^{#1}}
\newcommand{\OURSst}[1]{\text{OURS}_{\text{st}}^{#1}}

\begin{document}

\title{Better Patch Stitching for Parametric Surface Reconstruction}

\author{\vspace{0.5em}
	{Zhantao Deng \quad Jan Bedna\v{r}\'{i}k \quad Mathieu Salzmann \quad Pascal Fua} \\
	{CVLab, EPFL, Switzerland} \\
	{\small \{firstname.lastname\}@epfl.ch}\\
}

\maketitle

\begin{abstract}
Recently, parametric mappings have emerged as highly effective surface representations, yielding low reconstruction error. In particular, the latest works represent the target shape as an atlas of multiple mappings, which can closely encode object parts. Atlas representations, however, suffer from one major drawback: The individual mappings are not guaranteed to be consistent, which results in holes in the reconstructed shape or in jagged surface areas.

We introduce an approach that explicitly encourages global consistency of the local mappings. To this end, we introduce two novel loss terms. The first term exploits the surface normals and requires that they remain locally consistent when estimated within and across the individual mappings. The second term further encourages better spatial configuration of the mappings by minimizing novel \emph{stitching error}. We show on standard benchmarks that the use of normal consistency requirement outperforms the baselines quantitatively while enforcing better stitching leads to much better visual quality of the reconstructed objects as compared to the state-of-the-art.

\end{abstract}
\section{Introduction} \label{sec:introduction}

While multiple shape representations have been studied for deep learning based generative 3D surface reconstruction, learned parametric mappings~\cite{Yang18a} have emerged as highly competitive. They not only achieve the highest reconstruction accuracy, as measured by the Chamfer distance (CD), but also bring other important benefits, such as inherent support for arbitrary shape resolution without memory impact or access to higher order analytical surface properties~\cite{Bednarik20}.

In this context, the current standard consists of using an ensemble of parametric mappings, commonly referred to as \emph{patches}~\cite{Groueix18}, the union of which models the target surface. The inherent differentiability of neural networks can then be exploited to explicitly control the distortion of the patches, limit the undesirable patch overlap and yield better reconstructions in terms of the normal estimation~\cite{Bednarik20}.
Nevertheless, while this reduces the reconstruction error to a few millimeters on the standard ShapeNet benchmark~\cite{Chang15}, the generated shapes still suffer from severe artifacts. Most notably, the surfaces may include holes arising from unconnected patches or jagged areas where multiple patches overlap and/or intersect. Even though these phenomena have little impact on the quantitative metrics, such as CD or normal angular error, they significantly affect the visual quality of the generated objects. As such, they are ill-suited for direct use in the real-world downstream tasks, such as high-quality rendering, virtual reality or animation.

\begin{figure}[t]
\centering
  \centerline{\includegraphics[width=8.5cm]{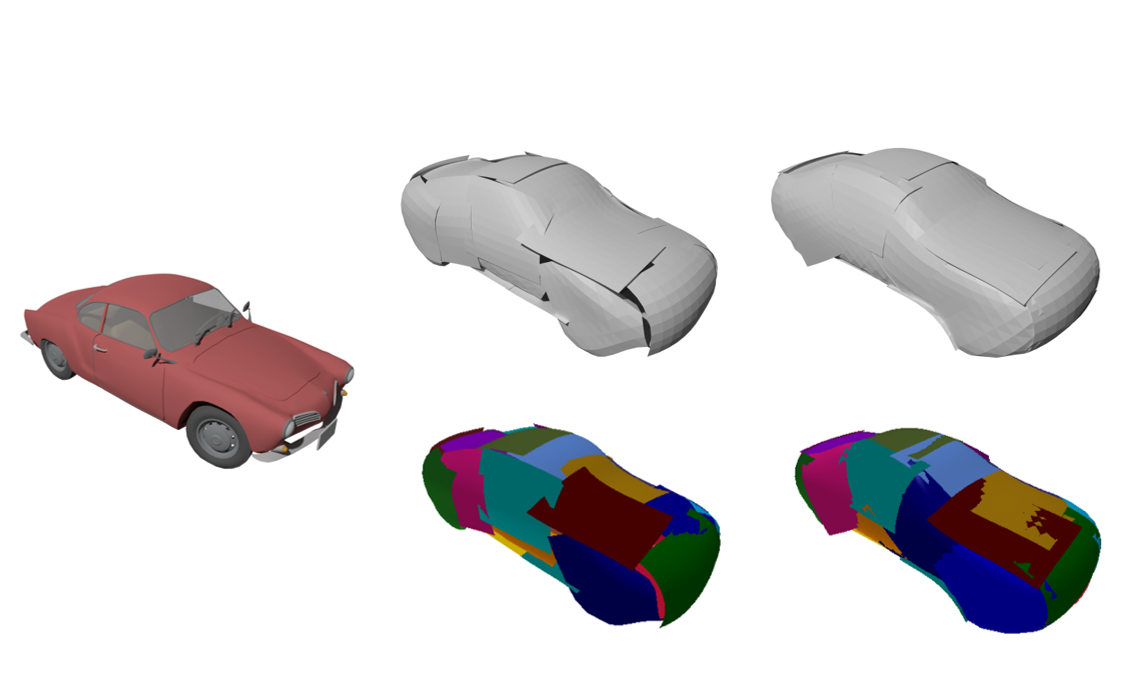}}
  \caption{\textbf{Qualitative comparison of our method with the SotA  approach (DSP) \cite{Bednarik20}}. From left to right: ground truth, DSP and our method. Different colors represent individual patches. The color bleeding (e.g. yellow and brown patch on the bottom right car) indicate that the given patches nearly coincide which is a sign of good stitching.}
\label{fig:teaser}
\end{figure}

In this paper, to address this, we observe that none of the existing learning and patch-based approaches~\cite{Groueix18,Deprelle19,Bednarik20} explicitly encode dependencies between the patches, which may ultimately lead to global inconsistencies. Therefore, we propose to explicitly encourage global surface consistency by linking each patch to its neighbors. In particular, we argue that correct estimation of higher-order surface properties, namely normals, has an impact on the surface reconstruction accuracy as measured quantitatively and thus should improve the visual quality. 

Therefore, we alter the training scheme so as to encourage normal consistency across the individual patches. Since surface normals correspond to the 1st order derivative, we effectively drive the learned mapping towards $C^{1}$ smooth function. Interestingly, while this loss term does indeed lead to higher accuracy reconstructions as shown in Section \ref{sec:experiments}, the visual quality is not always dramatically improved. Since we observe that, besides other defects, it is mainly surface holes causing undesirable visual artifacts, we couple the normal consistency requirement with additional new loss term designed to close the holes. As expected, this yields much better visual quality of the reconstructed surfaces while taking a small toll on the quantitative measurements. 

Our approach only relies on novel loss functions and thus can easily be incorporated into any existing patch-wise surface representation, regardless of the downstream task at hand. In particular, we deploy it within the state-of-the-art approach~\cite{Bednarik20}. This allows us to preserve the benefits of learning low-distortion mappings while producing surface reconstructions of higher quantitative precision and/or visual quality, as illustrated by Fig.~\ref{fig:teaser}.

In short, we contribute novel loss functions enforcing the consistency of estimated surface normals across the patches, as well as better spatial configuration of the mappings closing the surface holes. Our method yields reconstructions of higher visual quality and state-of-the-art results in terms of surface normals. We will make our code publicly available.

\section{Related Work} \label{sec:related_work}

\paragraph{3D Shape Representations.}
The field of deep learning based 3D shape reconstruction offers a rich palette of shape representations. Currently, the most popular ones are dense pixel-wise depth maps or normal maps~\cite{Eigen14,Bansal16,Bednarik18,Tsoli19,Zeng19}, point clouds~\cite{Fan17,Qi17a,Qi17b,Yang18a}, meshes~\cite{Wang18,Gundogdu19,Yao20,Yifan20}, implicit functions~\cite{Chen19,Mescheder19,Park19,Xu19,Atzmon20}, voxels~\cite{Choy16,Hane17}, shape primitives~\cite{Chen20,Deng20,Smirnov20,Paschalidou20}, parametric mappings~\cite{Yang18a,Groueix18,Williams19,Deprelle19,Bednarik20,Pang20,Badki20} or combinations of some of these~\cite{Muralikrishnan19,Poursaeed20}. The reason for the existence of so many representations is the fact that there is no silver bullet and each of them holds its specific benefits and drawbacks, often in terms of the trade-off between memory requirements, surface fitting precision, adaptability to different shape topologies or access to the surface properties. 

Furthermore, there exists a discrepancy between the metrics gauging the reconstruction quantitatively and perceived visual quality \cite{Wang18,Gkioxari19}. In particular, the work based on implicit surfaces \cite{Chen19,Mescheder19,Park19,Xu19} produces visually appealing reconstructions which, however, yield worse results as measured by CD than the standard parametric mapping method AtlasNet \cite{Groueix18}. We observe a similar phenomenon in our work, namely that significantly increased visual quality is often traded-off for lower quantitative precision.

We will focus on the category of parametric mappings, otherwise referred to as \emph{patch-wise} representations. It yields high surface fitting accuracy, has a low memory footprint, and provides access to analytical surface properties. However, it is known to suffer from inconsistencies across the individual patches~\cite{Groueix18,Williams19,Badki20}, which is the problem we tackle in this paper.

\vspace{-0.3cm}
\paragraph{Patch-wise Representations.}
FoldingNet~\cite{Yang18a} was the first deep net based work to model the surface of an object as a single learned parametric function embedding a 2D manifold into 3D space. Follow-up works then shifted to ensembles of such learned functions, the patch-wise representation, either in a learning ~\cite{Groueix18,Deprelle19,Bednarik20,Pang20} or an optimization~\cite{Williams19,Badki20} setting. A similar idea was recently used in the 2D output domain~\cite{Smirnov20}.

The popularity of patch-wise representations is typically due to their high precision in surface fitting. However, they are not yet flawless. One major drawback is that the learned mappings largely overlap and suffer from high distortion, which can only be alleviated by appropriate regularization~\cite{Bednarik20}. An even more pressing issue is the global inconsistency in the placement of the individual patches, which yields surface artifacts, such as holes or areas where multiple patches incorrectly intersect. This problem has been attacked~\cite{Williams19,Badki20} but only in the optimization setting, which is slow and requires access to geometrical observations (e.g., a noisy point cloud) at test time, and thus cannot be used for single view reconstruction. Our work builds on the learning based approach~\cite{Bednarik20} so as to retain its low-distortion and low-overlap properties while improving global patch consistency.

\vspace{-0.3cm}
\paragraph{Surface Distortion.}
When not explicitly constrained, the individual patches undergo large distortions, which is undesirable if one needs to compute analytical surface properties, such as normals or curvature, or when the mapping is to be used, e.g., for texture mapping. The distortion can be minimized either as a post-processing step~\cite{Groueix18} or penalized already at training time~\cite{Bednarik20}. 

Other applications where the surface is modeled as a mesh, i.e., a discrete approximate variant of the patch-wise representation, require control over distortion. To address this, \cite{Yifan20} introduced learned mesh cages for preserving surface details during object deformation, \cite{Cosmo20} learns a latent shape space via interpolation while imposing near-isometric deformation, \cite{Schmidt20} relies on finding correspondences between pairs of deformed objects based on a constant curvature metric assumption, and \cite{Tang20} seeks consistent UV mappings for better performance of the H.264 video codec.

\vspace{-0.3cm}
\paragraph{Estimation of Higher-order Surface Properties.}
When point clouds are involved, higher-order surface properties, such as normals and curvature, have been traditionally estimated via eigen-decomposition of the covariance matrix of a small point neighborhood \cite{Rusu09}. In the deep net setting, they can be predicted directly by the network~\cite{Groueix18}, computed analytically~\cite{Bednarik20} or estimated by iteratively adjusting an anisotropic kernel for weighted least squares plane fitting~\cite{Lenssen20}. Our approach exploits normals to drive the learned mappings towards $C^{1}$ smooth functions. We choose to compute this quantity analytically as it yields better performance than its approximate counterpart, as shown by our experiments in Section~\ref{sec:experiments}.

\section{Minimum Distortion Patch-wise Mapping} \label{sec:minimum_distortion_patch-wise_mapping}
For the sake of completeness, let us start by briefly formalizing the patch-wise representation~\cite{Groueix18} with distortion regularization~\cite{Bednarik20}, which we adopt in our approach. For more detail on this framework, we refer the reader to~\cite{Groueix18,Bednarik20}.

Let us assume a patch-wise representation with $K$ patches. The $k$-th patch is defined by a function $\fwk{k}: \real^{D} \times \real^{2} \rightarrow \real^{3}$, which maps a latent vector $\cb \in \real^{D}$ representing a shape and a point $\qb_{k_{i}} \in \real^{2}$ sampled from a 2D space $[0, 1]^{2}$ to a 3D point $\pb_{k_{i}} \in \real^{3}$. Each $\fwk{k}$ is modeled as a multi-layer perceptron (MLP) with its own set of trainable parameters $\wb_{k}$. Finally, the target shape is modeled as $\bigcup_{k}\fwk{k}$. Given $N$ ground-truth (GT) points $\pbhj$ and $M$ points $\pbik$ predicted for each patch $k$, the MLPs are trained jointly using the CD loss
\begin{small}
\begin{align}
\begin{split}
L_{CHD} = \sumKM&\min_{j}\left\Vert \pbik - \pbhj\right\Vert^2 + \\ 
\frac{1}{N}\sum_{j=1}^{N}&\min_{i,k}\left\Vert \pbik - \pbhj\right\Vert^2. \label{eq:chamfer}
\end{split}
\end{align}
\end{small}

Since the CD loss does not prevent the individual patches from collapsing, overlapping or undergoing severe distortions, additional regularization terms are exploited. Specifically, let the metric tensor computed for point $\pbik$ be defined as
\begin{equation}
    \gik = \begin{bmatrix}\Eik & \Fik \\ \Fik & \Gik \end{bmatrix},
\end{equation}
where $\Eik$, $\Fik$ and $\Gik$ encode how the 3D coordinates of $\pbik$ change with respect to a change in $\qb_{k_{i}}$.
Following~\cite{Bednarik20}, we define distortion and overlap regularizers as
\begin{align}
	\mathcal{L}_{[E|G]} &= \sumKM\left(\frac{[E|G]_{i}^{(k)} - \mu_{[E|G]}}{\Ak}\right)^{2}, \\
	\Lskew &= \sumKM\left(\frac{\Fik}{\Ak}\right)^{2}, \\
	\Loverlap &= \max\left(0, \sum_{k=1}^{K}\left(\Ak\right) - \hat{A}\right)^{2}, \label{eq:l_overlap}
\end{align}
where $\mu_{[E|G]}~=~\frac{1}{KM}\sum_{k}\sum_{i}[E|G]_{i}^{(k)}$, $\Ak$ is the predicted area of the $k$-th patch and $\hat{A}$ is the GT area of the target object. In short, these regularizers encourage that each patch remains rectangular, without local stretching or shrinking, and penalize the patches from overlapping each other. 

The overall loss function can then be written as
\begin{equation}
    \Ldsp = \Lchd + \alphE\LE + \alphG\LG + \alphskew\Lskew + \alpholap\Loverlap\; , \label{eq:defLoss}
\end{equation} 
where $\alphE, \alphG, \alphskew$ and $\alpholap$ are scalars used to weigh the individual terms. These loss terms, however, do not prevent holes or jagged areas from appearing between neighboring patches. This is what we tackle with the two new loss terms introduced in the following section.

\section{Methodology} \label{sec:methodology}

The goal of our approach is to prevent globally-inconsistent spatial configurations of the  patches, or in other words, incorrect patch \emph{stitching}. Below, we first identify the main stitching defects and then define two novel loss terms to counteract them.

\subsection{Limitations of Patch-Wise Representations} \label{ssec:limitations_of_patch_wise_representations}

Even though the loss function of Eq.~\ref{eq:defLoss} drives the patches towards the target surface while minimizing distortions and overlaps, it does not enforce correct patch stitching, which manifests itself through two main symptoms.

\begin{figure}[t]
\centering
\centerline{\includegraphics[width=8.5cm]{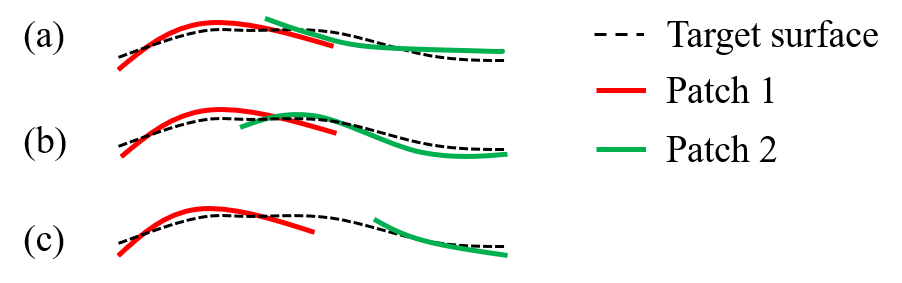}}
\caption{\textbf{1D representation of typical stitching problems.} (a) Overlaying patches, (b) crossing patches and (c) a hole between patches.}
\label{fig:patch_stitching_problems_schematic}
\end{figure}

\begin{figure}[t]
\centering
\centerline{\includegraphics[width=8.5cm]{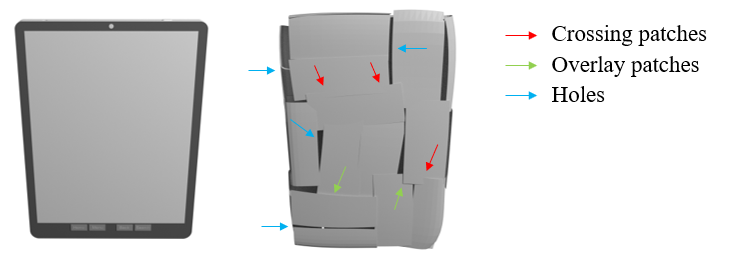}}
  \caption{\textbf{Undesirable visual artifacts caused by the typical stitching problems.} Overlays, crossings and holes.}
\label{fig:patch_stitching_problems_real}
\end{figure}

\paragraph{Crossings and Overlays.}
The loss term $\Loverlap$ of Eq.~\ref{eq:l_overlap} only affects the total cumulative area of all $k$ predicted patches, and thus does not prevent some of them to partially overlap. In practice, multiple patches end up fighting over the explanation of the true underlying surface, forming \emph{crossings} or \emph{overlays}, as depicted by Fig.~\ref{fig:patch_stitching_problems_schematic}(a,b). This issue is not necessarily reflected in the quantitative metrics, such as CD, but it causes undesirable visual artifacts, such as the ones shown in Fig.~\ref{fig:patch_stitching_problems_real}.

\paragraph{Holes.}
We consider holes to be the areas of the GT surface that are not covered by any of the predicted patches, as illustrated in Fig.~\ref{fig:patch_stitching_problems_schematic}(c). Depending on the size of the holes, this problem might affect the quantitative metrics, but it is typically more obvious qualitatively, as demonstrated in Fig.~\ref{fig:patch_stitching_problems_real}.

\subsection{Surface Consistency}\label{subsec:SurfConsis}
\begin{figure}[t]
    \centering
    \centerline{\includegraphics[width=0.48\textwidth]{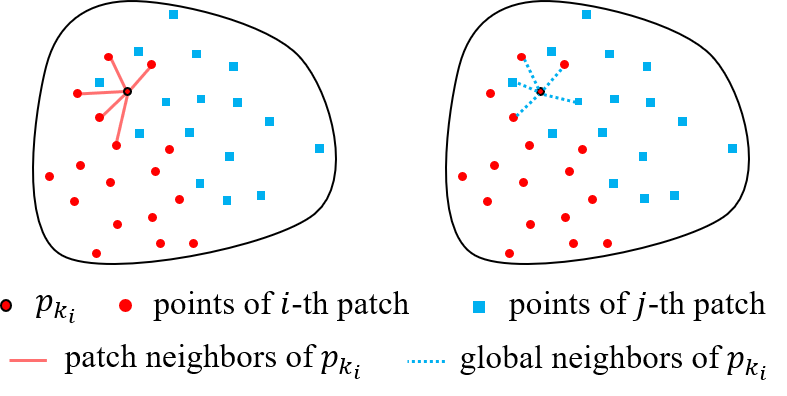}}
    \caption{\textbf{2D representation of the patch and global neighborhoods for point $\pbik$ in the overlapping part.}}
    \label{fig:neighborPoints_and_stitching}
\end{figure}

Let us now introduce our approach to addressing these two types of inconsistencies. To this end,
let $P_{k}, P_{l}$ be two partially overlapping patches, and $\pbik$ a point in $P_{k}$ spatially close to some points of patch $P_{l}$, as depicted by Fig.~\ref{fig:neighborPoints_and_stitching}. Given a set of points in the neighborhood of $\pbik$, one can estimate the normal vector of patch $P_{k}$ at $\pbik$.

We consider two types of a neighborhood of a point $\pbik$: A \emph{patch neighborhood}, consisting of the $n$ nearest points coming only from the same patch $P_{k}$; and a \emph{global neighborhood}, consisting of the $n$ nearest points coming from any patch (including $P_{k}$ itself), as visualized in Fig.~\ref{fig:neighborPoints_and_stitching}. We will denote a normal estimated using a patch neighborhood as $\nbkip$, and its counterpart estimated using a global neighborhood as $\nbkig$.

If the two patches $P_{k}$ and $P_{l}$ stitch well, we should have $\nbkip \approx \nbkig$. However, the stitching defects discussed in Section~\ref{ssec:limitations_of_patch_wise_representations} break these relations. We therefore propose to overcome these issues by enforcing consistency between the normals estimated using patch and global neighborhoods. 

To this end, we introduce a new loss function, which we dub \emph{surface consistency}, expressed as
\begin{equation}
    \Lsc = \sumKM  1 - ({\nbkip}^{\top}\nbkig)^{2}.
\end{equation}

This loss effectively drives the overlapping patches to coincide rather than overlay or crossover and stitches together the sufficiently small holes.

In practice, we estimate the normals using a standard covariance-based technique~\cite{Rusu09}. Note, however, that, as shown in~\cite{Bednarik20}, $\nbkip$ can be computed analytically through the network, which reduces the computation cost. We therefore experiment with both variants, and show in Section~\ref{sec:experiments} that the analytically computed normals perform better in our context. Note, however, that the normals estimated from the global neighborhood still have to be computed approximately using the covariance-based method so that the prospective inconsistencies would manifest.

\begin{figure}[t]
\centering
  \centerline{\includegraphics[width=0.48\textwidth]{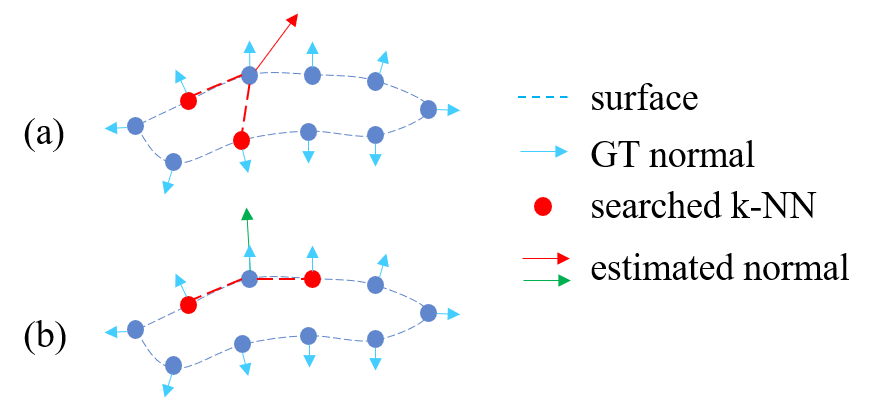}}
  \caption{\textbf{An example of the necessity to constrain the nearest neighbors search to the points with similar GT normal.} (a)~Unconstrained nearest neighbors could find points on the opposite side of the shape, leading to incorrect normal estimation; (b)~constrained nearest neighbors find valid points.}
\label{fig:constrained_knn}
\end{figure}

\paragraph{Constrained $n$ nearest neighbours.} A naive approach to obtaining the global neighborhood consists of performing a standard $n$ nearest-neighbor search. However, this is poorly suited to relatively thin object parts, such as the body of a cellphone or the top of a desk, because two opposite surfaces are spatially close to each other. The resulting neighborhood would then mix points from both surfaces, yielding highly inaccurate normals, as shown in~Fig.~\ref{fig:constrained_knn}.

To prevent this, we therefore consider a point $\pb_{j}$ to be a neighbor candidate of a point $\pb_{i}$ only if $\arccos(\hat{\nb_{i}}^{\top}\hat{\nb_{j}}) < \theta$, where $\hat{\nb_{i}}, \hat{\nb_{j}}$ are the GT normals of the GT points closest to $\pb_{i}$ and $\pb_{j}$, respectively, and $\theta$ is a threshold.

\subsection{Patch stitching error}\label{subsec:patchStit}
We observe that the loss term $\Lsc$, being dependent on $n$ nearest neighbors search, can successfully close only sufficiently small holes in the surface where the neighbors on the other side of a hole can be found. In practice, however, the reconstruction often suffers from bigger holes which would be left uncorrected by $\Lsc$. While this phenomenon does not manifest in the quantitative results strongly, visual quality is largely impaired by surface holes, as visualized in Fig.~\ref{fig:patch_stitching_problems_real}. Therefore, we formulate another loss term, which we call \emph{stitching error}, in order to counteract such problem.

\begin{figure}[htb]
\centering
\centerline{\includegraphics[width=0.49\textwidth]{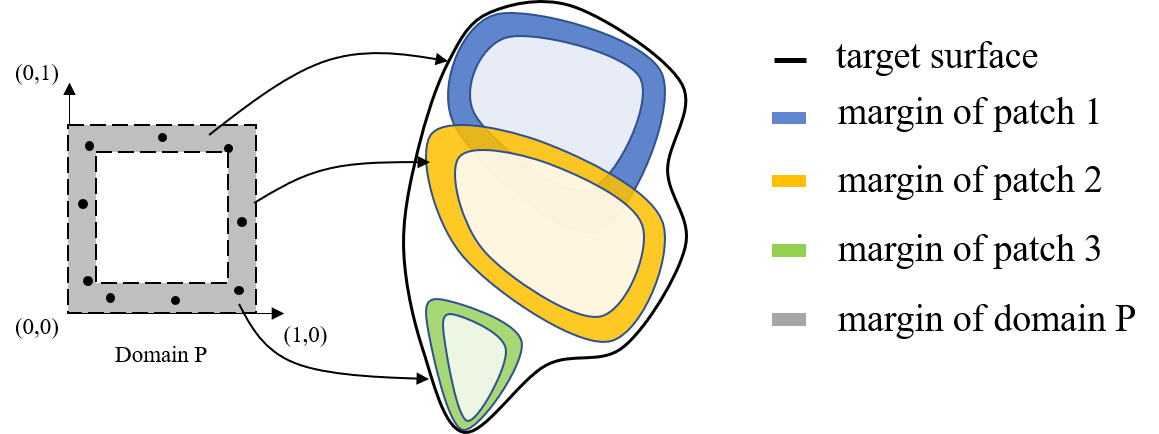}}
\caption{\textbf{2D representation of the overlaps and holes.} The margins of the patches are generated from the corresponding margins of the parameter space $P$, which is a unit square in our case.}
\label{fig:stitchingHoles}
\end{figure}

We start from the conjecture that in order to achieve satisfactory patch stitching, one only needs to consider the margins of a patch. In a desirable configuration, the margins of one patch would always coincide with the margins (or other parts) of other patches, as schematically shown by the blue and yellow patch in Fig. \ref{fig:stitchingHoles}. We thus construct a loss term which enforces that every margin of every patch must be spatially close in L2 sense to any other patch.

Let us define the patch margin in terms of its 2D parameter space. Let $r \in [0, 0.5]$ be the size of the 2D parameter space margin of patch $P_{k}$ which is in turn defined as a set of 2D points 
\begin{equation}
    P = \{(u, v) | u < r \lor u > 1 - r \lor v < r \lor v > 1 - r \},
\end{equation}
where $u, v \in [0, 1]$ are the coordinates of the 2D parameter space. Finally, the patch margin is a set of points $\{\pb_{k_{i}}^{m} | \pb_{k_{i}}^{m} = \fwk{k}(\cb, \qb_{k_{i}}^{m}) \land \qb_{k_{i}}^{m} \in P \}$.

Let $M_{k}$ be a number of points in the margin of patch $P_{k}$ and $\pb_{l_{j}}$ be the $j$-th 3D point of patch $P_{l}$. The stitching error is then defined as

\begin{equation}
    \Lst = \sum_{k=1}^{K}\frac{1}{M_k}\sum_{i=1}^{M_k}\min_{l \neq k}\min_{j}\left\Vert \pb_{k_{i}}^{m} - \pb_{l_{j}}\right\Vert^2. \label{eq:lst}
\end{equation}

\subsection{Final Loss Function}\label{subsec:final_loss_function}

We define our final combined loss function as

\begin{equation}
    \Lmain = \Ldsp + \alphsc\Lsc + \alphst\Lst, \label{eq:final_loss}
\end{equation}

where $\alphsc$ and $\alphst$ are scalar weights for the individual loss terms. In particular, as we show in Sec. \ref{sec:experiments}, the $\Lst$ term trades-off higher visual surface quality for lower reconstruction accuracy. Therefore, depending on the task at hand, it can be turned off by setting $\alphst = 0$.

\section{Experiments} \label{sec:experiments}

Following~\cite{Bednarik20}, we show that our approach can be used for various down-stream tasks. In particular, we present results on point cloud auto-encoding (\PCAE{}) and single-view reconstruction (\SVR{}). 

Since our approach extends the loss function of Eq.~\ref{eq:defLoss} with two novel loss terms, we investigate their individual contribution, including the possibility of removing some of the original terms of Eq.~\ref{eq:defLoss}. Finally, we show quantitative and qualitative comparisons to the SotA~\cite{Bednarik20}.

\subsection{Tasks, Data and Architectures} \label{sec:task_data_arch}

The goal of the \PCAE{} task is to encode the input point cloud into a compact global descriptor, which is then used to generate the original shape. By contrast, the \SVR{} task aims at reconstructing the 3D shape of an object observed in an input RGB image. In both cases, we use exactly the same architectures as in~\cite{Bednarik20}, that is, an auto-encoder (AE) structure whose decoder is 4-layer MLP with Softplus activation functions, and whose encoder is a PointNet~\cite{Qi17a} for \PCAE{} and a ResNet18~\cite{He16} for \SVR{}.

Both tasks are evaluated on the standard benchmark dataset ShapeNet Core v2~\cite{Chang15}. This synthetic dataset consists of multiple categories of CAD objects and is commonly used to benchmark 3D reconstruction tasks. The dataset splits used in~\cite{Bednarik20} only contain training and testing subsets. Since we need to perform grid search for $n$, we split the testing subset into two halves, the first one is used for validation and the second one for testing.

Following~\cite{Bednarik20}, we set the loss term weights of Eq.~\ref{eq:defLoss} to $\alphE = \alphG = \alphskew = 0.001$ and $\alpholap = 0.1$ in all our experiments. 
We set the weights of Eq.~\ref{eq:final_loss} to $\alphsc = 0.001$ and $\alphst = 0.001$ or $\alphst = 0$ depending on the model variant at test. The margin is set to $r = 0.1$, the angular threshold for the constrained $n$ nearest neighbors to $\theta = 120^{\circ}$, and $n$ is chosen individually in each experiment using grid search on the validation set.

We observed that starting the training from scratch with our novel loss terms not only does not improve the results but can even slow down and/or harm the training. The reason is that both $\Lsc$ and $\Lst$ require that the prediction is already sufficiently close to the correct solution. However, at the beginning of the training the network predicts patches with random positions and scale. Therefore, in all our experiments, we first pretrain the network before adding $\Lsc$ and/or $\Lst$. Note, however, that all the training runs including the SotA work which we compare with, are trained for exactly the same total number of iterations for fair comparison.

\subsection{Metrics}

We use the following metrics to report the performance of all the tested methods.

\parag{Chamfer distance (CD).} The formula is given in Eq.~\ref{eq:chamfer}.

\parag{Angular error ($\mathbf{\mAE}$).} This metric is defined as $\mAE = \frac{1}{M}\sum_{i=1}^{M} \arccos{|\nb_{i}\hat{\nb_{i}}|}$, where $\nb_{i}$ is the analytically-computed normal of the predicted point $\pbi$, and $\hat{\nb_{i}}$ is the GT normal of the GT point closest to $\pbi$. The values are reported in degrees.

\parag{Stitching metric ($\mathbf{\mS}$).} We use the loss defined in Eq.~\ref{eq:lst} also as a metric both to gauge the performance of the method using only the $\Lsc$ term and also to show the extent to which optimizing directly for $\Lst$ can minimize this quantity.

\parag{Overlap ($\mathbf{\molap^{(t)}}$).} The overlap is defined in~\cite{Bednarik20} as the number of patches within a distance $t$ from a given predicted point, averaged over all the points.
\\~

As opposed to \cite{Bednarik20}, we do not report the amount of collapsed patches because, thanks to the distortion regularizers, the patches do not collapse.

\subsection{Ablation Study} \label{sec:ablation_study}

In this section, we study the influence of the different loss terms on the results of our approach. Beside removing one or both of our novel terms $\Lsc, \Lst$, we also experiment with removing the term $\Loverlap$, which we conjecture might be the main cause for the undesirable surface holes. Furthermore, as we stated in Section~\ref{subsec:SurfConsis}, we compare the results of our approach obtained with per-point normals computed analytically vs approximately using point neighborhoods. In all these experiments, we use neighborhoods of $n = 8$ points as justified by the grid search which we discuss in detail in Section \ref{ssec:pcae_and_svr}.

We perform our ablation studies on the \PCAE{} task for multiple objects of the ShapeNet dataset. Since we observed similar trends for all objects, for brevity we report the results on the \textbf{plane} object category only, but provide the exhaustive results in the supplementary material. We refer to the different variants of our approach as follows:
\begin{itemize}
    \item \textbf{\OURSaprx}: approximate normals, only $\Lsc$.
    \item \textbf{\OURSanlt}: analytical normals, only $\Lsc$.
    \item \textbf{\OURSstitch}: without $\Lsc$, only $\Lst$.
    \item \textbf{\OURSanltstch}: analytical normals, $\Lsc$ and $\Lst$.
    \item \textbf{\OURSanltarea}: analytical normals, only $\Lsc$, no $\Loverlap$.
    \item \textbf{\OURSanltstcharea}: analytical normals, $\Lsc$ and $\Lst$, no $\Loverlap$.
\end{itemize}
For reference, we also include the results of the approach of~\cite{Bednarik20}, which we refer to as \DSP{}. 

The results are provided in Table~\ref{tab:ablationStudy}. We observe that, in general, analytically computed normals lead to better results in all metrics, and therefore use them in all the following experiments. Removing the $\Loverlap$ loss term does not lead to consistently better performance, and thus we keep this term in our next experiments. Finally, the $\Lsc$ term clearly leads to higher accuracy in terms of angular error $\mAE$ and stitching metric $\mS$, while slightly degrading the CD metric. Adding  $\Lst$ to $\Lsc$ dramatically improves the $\mS$ metric, while sacrificing some precision in all the other metrics.

 \begin{table}[t]
  \centering
  \caption{Ablation study on the \textbf{plane} category of ShapeNet. All models are based on the same pretrained model. The CD is multiplied by 1e3.}
    \begin{tabular}{ccccc}
    \textbf{Method} & \textbf{CD} & $\mathbf{\mAE}$ & $\mathbf{\mS}$ & $\mathbf{\molap^{(0.05)}}$ \\
    \midrule
    \textbf{\DSP{}} & \textbf{1.23} & 18.96 & 0.854 & \textbf{2.41} \\
    \textbf{\OURSaprx} & 1.27 & 18.85 & 0.799 & 2.74 \\
    \textbf{\OURSanlt} & 1.26  & 18.62 & 0.796 & 2.62 \\
    \textbf{\OURSstitch} & 1.30 & 19.71 & 0.468 & 3.07 \\
    \textbf{\OURSanltstch} & 1.33 & 18.85 & \textbf{0.456} & 3.01 \\
    \textbf{\OURSanltarea} & 1.26  & \textbf{18.56} & 0.793 & 2.60 \\
    \textbf{\OURSanltstcharea} & 1.42  & 19.94 & 0.474 & 3.29 \\
    \bottomrule
    \end{tabular}
  \label{tab:ablationStudy}
\end{table}

\subsection{PCAE and SVR} \label{ssec:pcae_and_svr}
In this section we compare our approach against the SotA \cite{Bednarik20} (DSP) on two tasks, PCAE and SVR, on the standard benchmark dataset ShapeNet. Following \cite{Bednarik20}, we use object categories cellphone, chair, couch, plane and car. For all the experiments we consider two variants of our approach: $\OURS{n}$ which only uses the surface consistency term $\Lsc$ (i.e. we set $\alphst = 0$) and $\OURSst{n}$ which combines the $\Lsc$ with the stitching loss term $\Lst$. The $n$ denotes the size of the global neigborhood used for the constrained $n$ nearest neighbors search as explined in Section \ref{subsec:SurfConsis}. 

The normal estimate is sensitive to the choice of $n$, therefore we perform a grid search for every ShapeNet object category using the validation split. Each value of $n$ requires a full training of the neural network which is costly. As we observed that lower values of $n$ in general perform better, we only consider $n \in \{4, 8\}$. Whichever value performs better on the validation set is used to report the test results. We provide complete results of the grid search in the supplementary material. Note that even with such a limited grid search we are able to achieve state-of-the-art results, however more careful tuning might lead to even better results.

As justified in Section~\ref{sec:ablation_study}, we compute the local patch normals analytically. Furthermore, as explained in Section~\ref{sec:task_data_arch} all the experiments of the given task start from the same pretrained model and are trained for the same number of iterations for the sake of fair comparison.

\begin{figure*}[htb]
\label{fig:samples}
\centering
\centerline{\includegraphics[width=0.81\textwidth]{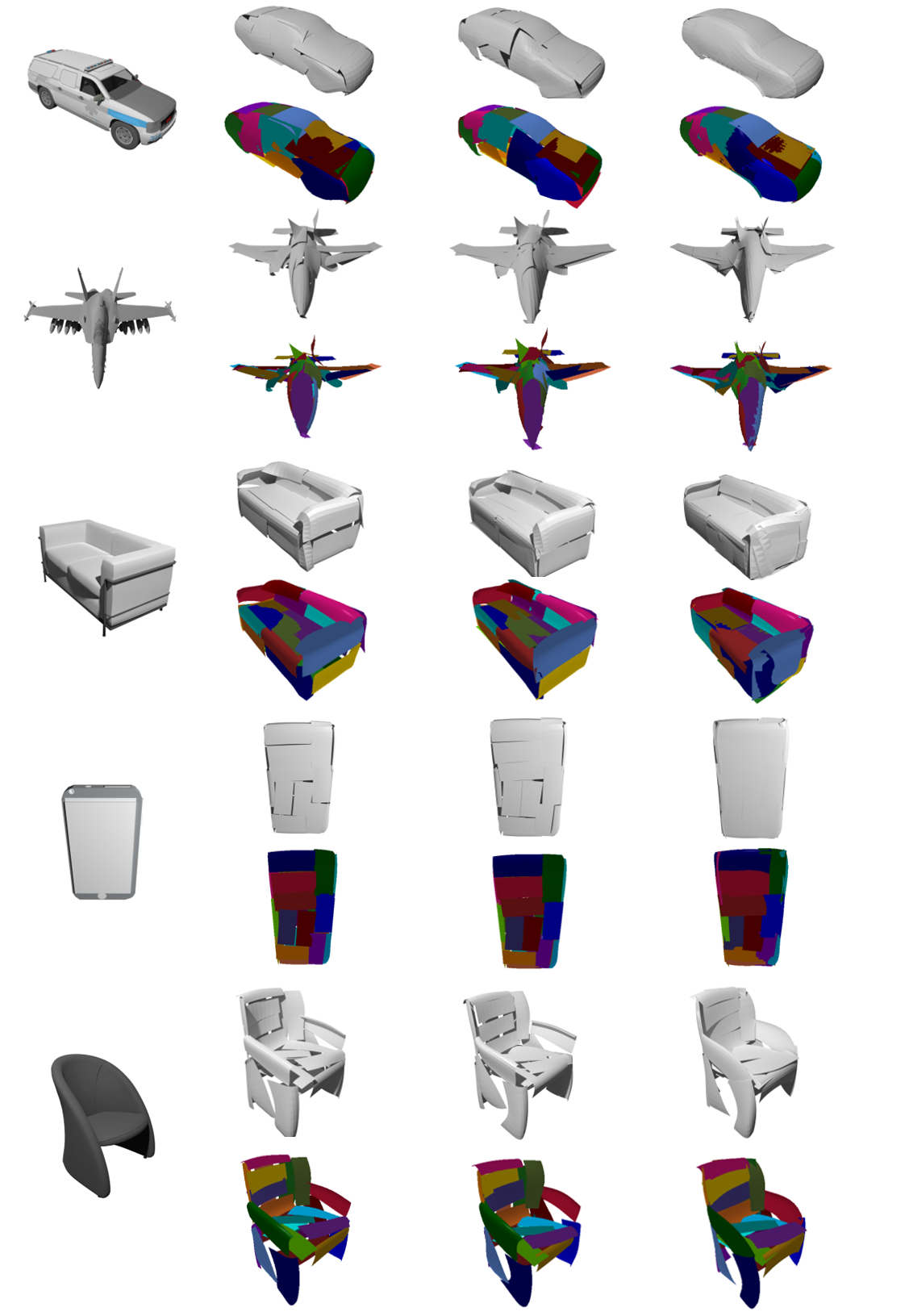}}
\caption{\textbf{Qualitative results of the \PCAE{} task.} From top to bottom: car, plane, couch, cellphone, chair. From left to right: ground truth, DSP, $\OURS{n}$ and $\OURSst{n}$. The choice of $n$ for each category corresponds to that of Table \ref{tab:OURS_VS_DSP}. The colors represent individual patches. The color bleeding visual artifact visible e.g. between the violet and blue patch on the nose of the right-most plane indicates that the patches nearly coincide which a sign of good stitching.}
\label{fig:final_results}
\end{figure*}

\paragraph{PCAE.} 
Following \cite{Bednarik20}, we train all the models with $25$ patches and use $2500$ points randomly sampled from the GT and the same amount for prediction. The quantitative results are reported in Table~\ref{tab:OURS_VS_DSP}, the qualitative ones in Fig.~\ref{fig:final_results}. 

$\OURS{n}$ consistently improves on normals with respect to DSP as measured by $\mAE$ and it yields lower error in terms of CD and $\mS$ in majority cases. Small gains in quantitative accuracy lead to only minor visual improvements, which can be observed mostly in the form of limiting some overlays and crossings (right wing of the airplane, hood of the car). However, introducing the stitching loss term $\Lst$ used in the model $\OURSst{n}$ brings a drastic decrease in the stitching metric $\mS$ which is traded off for an increase in angular error $\mAE$ and CD. Despite the lower quantitative score, $\OURSst{n}$ clearly offers the best visual quality of the reconstructed shapes which is mostly due to successfully closed surface holes.

\begin{table}[t]
  \begin{center}
    \caption{\textbf{Comparison of our methods with DSP on \PCAE{} task.} The CD is multiplied by 1e3.}
    \label{tab:OURS_VS_DSP}
    \begin{tabular}{c c c c c c}
      \textbf{Category} 
      & \textbf{Method} 
      & \textbf{CD} 
      & $\mathbf{\mAE}$ & $\mathbf{\mS}$ & $\mathbf{\molap^{(0.05)}}$ \\
      \hline 
                 & DSP & \textbf{2.05} & \textbf{12.55} & 0.648 & 1.76\\
      cellphone  & $\OURS{4}$ & 2.05 & 12.66 & 0.655 & \textbf{1.73}\\
                 & $\OURSst{4}$ & 2.07 & 13.06 & \textbf{0.578} & 2.14\\
      \hline
            & DSP & 3.05 & 25.67 & 1.69 & 1.70\\
      chair & $\OURS{4}$ & \textbf{2.96} & \textbf{25.50} & 1.71 & \textbf{1.63}\\
            & $\OURSst{4}$ & 3.96 & 29.33 & \textbf{1.09} & 1.79\\
      \hline
            & DSP & \textbf{2.17} & 17.25 & 0.807 & 1.95\\ 
      couch & $\OURS{4}$ & 2.20 & \textbf{17.07} & 0.806 & \textbf{1.86}\\
            & $\OURSst{4}$ & 2.36 & 18.68 & \textbf{0.692} & 2.21\\
      \hline
            & DSP & \textbf{1.23} & 18.96 & 0.854 &\textbf{ 2.41}\\
      plane & $\OURS{8}$ & 1.26 & \textbf{18.62} & 0.796 & 2.62\\
            & $\OURSst{8}$ & 1.33 & 18.85 & \textbf{0.456} & 3.01\\
      \hline
            & DSP & 1.72 & 18.45 & 0.762 & 1.88\\
      car   & $\OURS{4}$ & \textbf{1.68} & \textbf{17.82} & 0.730 & \textbf{1.79}\\
            & $\OURSst{4}$ & 1.79 & 18.22 & \textbf{0.566} & 2.10\\
      \hline
    \end{tabular}
  \end{center}
\end{table}

\vspace{-0.2cm}
\paragraph{SVR.}
As was the case for the \PCAE{} task, we again train all the models with $25$ patches, use $2500$ points sampled randomly from the GT and the same amount for prediction. We report the quantitative results in Table~\ref{tab:OURS_VS_DSP_on_SVR_SN}. Since the qualitative results follow a similar trend to that of the \PCAE{} task, we do not report them here but refer the reader to the supplementary material.

Quantitative results follow a similar trend as well. That is, $\OURS{n}$ consistently improves on normals with respect to DSP as measured by $\mAE$ and it yields lower error in terms of the stitching metric $\mS$ in majority cases. Again, the improvement of the visual quality is rather marginal. $\OURSst{n}$ brings again a large improvement in stitching metric $\mS$.  However, this time it not only improves stitching but also the angular error $\mAE$. We can attribute this to the fact, that \SVR{} is in general much harder task than \PCAE{} (as is clear from overall lower CD and $\mAE$ accuracy) and the reconstructions are thus already less precise to begin with. Enforcing better patch stitching as per $\Lst$ thus not only improves the visual quality of the surface but the reconstruction precision as well.

\begin{table}[t]
  \begin{center}
    \caption{\textbf{Comparison of our methods with DSP on \SVR{} task.} The CHD is multiplied by 1e3.}
    \label{tab:OURS_VS_DSP_on_SVR_SN}
    \begin{tabular}{c c c c c c}
      \textbf{Category} 
      & \textbf{Method} 
      & \textbf{CD} 
      & $\mathbf{\mAE}$ & $\mathbf{\mS}$ & $\mathbf{\molap^{(0.05)}}$ \\
      \hline 
                 & DSP & \textbf{8.44} & 13.52 & 1.39 & 1.43\\
      cellphone  & $\OURS{8}$ & 8.91 & \textbf{13.18} & 1.34 & \textbf{1.38}\\
                 & $\OURSst{8}$ & 8.63 & 16.10 & \textbf{0.926} & 1.47\\
      \hline
            & DSP & \textbf{7.07} & 38.85 & 1.79 & 1.52 \\
      chair & $\OURS{4}$ & 7.31 & 38.96 & 1.91 & \textbf{1.50}\\
            & $\OURSst{4}$ & 7.60 & \textbf{35.98} & \textbf{1.14} & 1.61\\
      \hline
            & DSP & 6.31 & 29.28 & 1.08 & 1.65\\ 
      couch & $\OURS{8}$ & \textbf{6.21} & 28.07 & \textbf{0.981} & 1.76\\
            & $\OURSst{8}$ & 7.51 & \textbf{24.75} & 1.12 & \textbf{1.43}\\
      \hline
            & DSP & \textbf{2.66} & 25.61 & 0.960 & \textbf{2.60}\\
      plane & $\OURS{4}$ & 2.72 & 23.95 & 0.908 & 2.61\\
            & $\OURSst{4}$ & 2.73 & \textbf{23.38} & \textbf{0.539} & 2.84\\
      \hline
            & DSP & 5.23 & 26.01 & 1.02 & \textbf{1.42}\\
      car   & $\OURS{8}$ & \textbf{3.46} & 24.01 & 0.830 & 1.60\\
            & $\OURSst{8}$ & 3.65 & \textbf{22.75} & \textbf{0.620} & 1.79\\
      \hline
    \end{tabular}
  \end{center}
\end{table}

\section{Conclusion} \label{sec:conclusion}
We have presented a new approach to 3D shape reconstruction which focuses on improving the quality of the reconstructed surfaces. We introduced two novel loss terms which attempt to encourage inter patch consistency in terms of the estimated surface normals and to further spatially bring the patches closer to each other, thus limiting the main defects of patch-wise approaches such as crossings, overlays or holes. Even though we strive to achieve both better qualitative and quantitative results, we showed that these two measures do not always necessarily go hand in hand and that one can achieve higher visual quality with lower numerical accuracy and vice versa. Depending on the loss term in use, we show that we can achieve SotA results in terms of angular error of the normals or in terms of visual quality.

As can be seen in the qualitative results, the quality is largely improved by our approach however it is not yet perfect for seamless use of the reconstructed objects as downstream tasks assets. Therefore, in the future work, we will explore the possibilities to improve the visual quality of the reconstructed surfaces even further. One promising direction might be to explore the current differentiable rendering approaches to directly minimize the visual discrepancies between the smooth GT and jagged reconstructed surfaces.

\paragraph{Acknowledgements} This work was supported in part by the Swiss National Science Foundation.

{\small
\bibliographystyle{ieee}
\bibliography{egbib}
}

\newpage
~\\
\newpage
\section{Supplementary Material}
We present the full ablation study performed on all five ShapeNet object categories selected for our experiments in Section \ref{sec:ablation}, complete results of the grid search for the best value of the hyperparameter $n$ of the constrained $n$ nearest neighbors in Section \ref{sec:gridSearch} and finally, qualitative results of the single-view reconstruction (\SVR{}) task in Section \ref{sec:qualitative_svr}.

\subsection{Ablation Study} \label{sec:ablation}
As explained in Section 5.3. of the paper, we study the influence of the individual loss terms of the Eq. 10. We performed the ablation study on the point cloud auto-encoding (\PCAE{}) task using five object categories of the ShapeNet dataset, namely cellphone, chair, couch, plane and car. For every object category, we set the hyperparameter $n$ of the constrained $n$ nearest neighbors to the value found by the grid search as detailed in Section \ref{sec:gridSearch}. The results are listed in Tables \ref{tab:ablationStudyCellphone}-\ref{tab:ablationStudyPlane}. As discussed in Section 5.3. of the paper, two main conclusions can be drawn from the results. 

First, removing the loss term $\Loverlap$ does not lead to consistently improved or consistently worse results. While \textbf{\OURSanltstcharea} in general achieves better results as compared to \textbf{\OURSanltstch} on most of the metrics, the same cannot be said about the comparisons of \textbf{\OURSanltarea} and \textbf{\OURSanlt} where the former in general yields better results in terms of the $\mS$ metric while the latter rather improves CD and $\mAE$. Therefore we decided to keep the $\Loverlap$ term to stay consistent with the method DSP \cite{Bednarik20} to which we compare our work. 

Second, while approximately computed normals (\textbf{\OURSaprx}) in general lead to an improvement as measured by the metric $\mS$, the analytically computed normals (\textbf{\OURSanlt}) consistently lead to lower angular error as measured by $\mAE$. As the analytical normals are also faster to compute, we chose to use them for all our experiments.

\begin{table}[ht]
  \centering
  \caption{Ablation study on the \textbf{cellphone} category of ShapeNet. All models are based on the same pretrained model. The CD is multiplied by 1e3.}
    \begin{tabular}{ccccc}
    \textbf{Method} & \textbf{CD} & $\mathbf{\mAE}$ & $\mathbf{\mS}$ & $\mathbf{\molap^{(0.05)}}$ \\
    \midrule
    \textbf{\DSP{}} & \textbf{2.05} & \textbf{12.55} & 0.648 & 1.76 \\
    \textbf{\OURSaprx} & 2.06  & 12.89 & 0.651 & 1.74 \\
    \textbf{\OURSanlt} & 2.05  & 12.66 & 0.655 & \textbf{1.73} \\
    \textbf{\OURSstitch} & 2.06 & 13.41 & 0.584 & 2.13 \\
    \textbf{\OURSanltstch} & 2.07  & 13.06 & 0.578 & 2.14 \\
    \textbf{\OURSanltarea} & 2.07 & 12.78 & 0.648 & 1.76\\
    \textbf{\OURSanltstcharea} & 2.05 & 12.66 & \textbf{0.570} & 2.27\\
    \bottomrule
    \end{tabular}
  \label{tab:ablationStudyCellphone}
\end{table}%

\begin{table}[t]
  \centering
  \caption{Ablation study on the \textbf{couch} category of ShapeNet. All models are based on the same pretrained model. The CD is multiplied by 1e3.}
    \begin{tabular}{ccccc}
    \textbf{Method} & \textbf{CD} & $\mathbf{\mAE}$ & $\mathbf{\mS}$ & $\mathbf{\molap^{(0.05)}}$ \\
    \midrule
    \textbf{\DSP{}} & \textbf{2.17} & 17.25 & 0.807 & 1.95 \\
    \textbf{\OURSaprx} & 2.22  & 17.30 & 0.797 & 1.87 \\
    \textbf{\OURSanlt} & 2.20  & 17.07 & 0.806 & \textbf{1.86} \\
    \textbf{\OURSstitch} & 2.32 & 18.05 & 0.724 & 2.16 \\
    \textbf{\OURSanltstch} & 2.36  & 18.68 & 0.692 & 2.21 \\
    \textbf{\OURSanltarea} & 2.20 & 17.48 & 0.804 & 1.88 \\
    \textbf{\OURSanltstcharea} & 2.25 & \textbf{17.02} & \textbf{0.677} & 2.46\\
    \bottomrule
    \end{tabular}
  \label{tab:ablationStudyCouch}
\end{table}%

\begin{table}[t]
  \centering
  \caption{Ablation study on the \textbf{chair} category of ShapeNet. All models are based on the same pretrained model. The CD is multiplied by 1e3.}
    \begin{tabular}{ccccc}
    \textbf{Method} & \textbf{CD} & $\mathbf{\mAE}$ & $\mathbf{\mS}$ & $\mathbf{\molap^{(0.05)}}$ \\
    \midrule
    \textbf{\DSP{}} & 3.05 & 25.67 & 1.69 & 1.70 \\
    \textbf{\OURSaprx} & 2.93  & \textbf{24.98} & 1.69 & 1.67 \\
    \textbf{\OURSanlt} & 2.96  & 25.50 & 1.71 & \textbf{1.63} \\
    \textbf{\OURSstitch} & 3.22 & 27.44 & 1.07 & 1.79 \\
    \textbf{\OURSanltstch} & 3.96  & 29.34 & 1.09 & 1.79 \\
    \textbf{\OURSanltarea} & \textbf{2.84} & 26.13 & 1.47 & 2.07 \\
    \textbf{\OURSanltstcharea} & 2.84 & 27.01 & \textbf{0.829} & 2.61 \\
    \bottomrule
    \end{tabular}
  \label{tab:ablationStudyChair}
\end{table}%

\begin{table}[t]
  \centering
  \caption{Ablation study on the \textbf{car} category of ShapeNet. All models are based on the same pretrained model. The CD is multiplied by 1e3.}
    \begin{tabular}{ccccc}
    \textbf{Method} & \textbf{CD} & $\mathbf{\mAE}$ & $\mathbf{\mS}$ & $\mathbf{\molap^{(0.05)}}$ \\
    \midrule
    \textbf{\DSP{}} & 1.72 & 18.45 & 0.762 & 1.88 \\
    \textbf{\OURSaprx} & 1.72  & 17.93 & 0.726 & 1.79 \\
    \textbf{\OURSanlt} & \textbf{1.68}  & 17.82 & 0.730 & \textbf{1.79} \\
    \textbf{\OURSstitch} & 1.83 & 19.02 & 0.593 & 2.02 \\
    \textbf{\OURSanltstch} & 1.79  & 18.22 & 0.566 & 2.10 \\
    \textbf{\OURSanltarea} & 1.72 & 17.65 & 0.746 & 1.82 \\
    \textbf{\OURSanltstcharea} & 1.79 & \textbf{17.25} & \textbf{0.544} & 2.57 \\
    \bottomrule
    \end{tabular}
  \label{tab:ablationStudyCar}
\end{table}%

\begin{table}[t]
  \centering
  \caption{Ablation study on the \textbf{plane} category of ShapeNet. All models are based on the same pretrained model. The CD is multiplied by 1e3.}
    \begin{tabular}{ccccc}
    \textbf{Method} & \textbf{CD} & $\mathbf{\mAE}$ & $\mathbf{\mS}$ & $\mathbf{\molap^{(0.05)}}$ \\
    \midrule
    \textbf{\DSP{}} & \textbf{1.23} & 18.96 & 0.854 & \textbf{2.41} \\
    \textbf{\OURSaprx} & 1.27 & 18.85 & 0.799 & 2.74 \\
    \textbf{\OURSanlt} & 1.26  & 18.62 & 0.796 & 2.62 \\
    \textbf{\OURSstitch} & 1.30 & 19.71 & 0.468 & 3.07 \\
    \textbf{\OURSanltstch} & 1.33 & 18.85 & \textbf{0.456} & 3.01 \\
    \textbf{\OURSanltarea} & 1.26  & \textbf{18.56} & 0.793 & 2.60 \\
    \textbf{\OURSanltstcharea} & 1.42  & 19.94 & 0.474 & 3.29 \\
    \bottomrule
    \end{tabular}
  \label{tab:ablationStudyPlane}
\end{table}

\subsection{Grid Search} \label{sec:gridSearch}
As discussed in Section 5.4. of the paper, the normal estimated via global neighborhood is sensitive to the choice of the hyperparameter $n$ of the constrained $n$ nearest neighbors search. Therefore we performed a grid search on every ShapeNet object category using the validation set. Due to the high computation cost of training the full neural network from scratch we only consider $n \in \{4, 8\}$. 

The results for the \PCAE{} and \SVR{} tasks are summarized in Table \ref{tab:gridSearch_PCAE} and Table \ref{tab:gridSearch_SVR} respectively. It can be clearly seen the within each object category, either $\OURS{4}$ and $\OURSst{4}$ or $\OURS{8}$ and $\OURSst{8}$ perform better in the majority of the metrics and thus that value of $n$ is used for evaluation on the test set.

\begin{table}[ht]
  \begin{center}
    \caption{\textbf{Grid search for the optimal $n$ on \PCAE{} task.} The CD is multiplied by 1e3.}
    \label{tab:gridSearch_PCAE}
    \begin{tabular}{c c c c c c}
      \textbf{Category} 
      & \textbf{Method} 
      & \textbf{CD} 
      & $\mathbf{\mAE}$ & $\mathbf{\mS}$ & $\mathbf{\molap^{(0.05)}}$ \\
      \hline 
                 & $\OURS{4}$ & \textbf{1.26} & \textbf{11.34} & 0.603 & \textbf{1.79}\\
      cellphone  & $\OURSst{4}$ & 1.30 & 11.75 & 0.547 & 2.22 \\
                 & $\OURS{8}$ & 1.27 & 12.15 & 0.590 & 1.82\\
                 & $\OURSst{8}$ & 1.31 & 12.26 & \textbf{0.542} & 2.22\\
      \hline
            & $\OURS{4}$ & \textbf{2.22} & \textbf{16.90} & 0.810 & \textbf{1.86}\\
      couch & $\OURSst{4}$ & 2.33 & 18.51 & \textbf{0.695} & 2.21\\
            & $\OURS{8}$ & 2.49 & 19.37 & 0.844 & 1.95\\
            & $\OURSst{8}$ & 2.34 & 17.82 & 0.708 & 2.21\\
      \hline
            & $\OURS{4}$ & \textbf{2.84} & \textbf{24.98} & 1.69 & \textbf{1.65}\\
      chair & $\OURSst{4}$ & 3.86 & 28.93 & \textbf{1.09} & 1.80\\
            & $\OURS{8}$ & 3.26 & 26.93 & 1.61 & 1.75\\
            & $\OURSst{8}$ & 6.85 & 31.11 & 1.24 & 1.91\\
      \hline
            & $\OURS{4}$ & 1.09 & 18.45  & 0.900 & \textbf{2.32}\\
      plane & $\OURSst{4}$ & 1.24 & 18.79 & 0.443 & 3.01\\
            & $\OURS{8}$ & \textbf{1.03} & \textbf{17.93} & 0.795 & 2.62\\
            & $\OURSst{8}$ & 1.12 & 18.11 & \textbf{0.438} & 3.02\\
      \hline
            & $\OURS{4}$ & 2.42 & \textbf{18.05} & 0.733 & 1.80\\
      car   & $\OURSst{4}$ & 2.46 & 18.45 & \textbf{0.574} & 2.11\\
            & $\OURS{8}$ & \textbf{2.34} & 18.28 & 0.787 & \textbf{1.69}\\
            & $\OURSst{8}$ & 2.61 & 18.79 & 0.580 & 2.08\\
      \hline
    \end{tabular}
  \end{center}
\end{table}

\begin{table}[t]
  \begin{center}
    \caption{\textbf{Grid search for the optimal $n$ on \textbf{SVR} task.} The CD is multiplied by 1e3.}
    \label{tab:gridSearch_SVR}
    \begin{tabular}{c c c c c c}
      \textbf{Category} 
      & \textbf{Method} 
      & \textbf{CD} 
      & $\mathbf{\mAE}$ & $\mathbf{\mS}$ & $\mathbf{\molap^{(0.05)}}$ \\
      \hline 
                 & $\OURS{4}$ & \textbf{6.49} & 15.53 & 1.27 & \textbf{1.36}\\
      cellphone  & $\OURSst{4}$ & 7.25 & 12.61 & 1.16 & 1.37 \\
                 & $\OURS{8}$ & 6.97 & \textbf{11.34} & 1.31 & 1.37\\
                 & $\OURSst{8}$ & 6.61 & 14.50 & \textbf{0.91} & 1.46\\
      \hline
            & $\OURS{4}$ & \textbf{7.59} & 38.62 & 1.90 & \textbf{1.50}\\
      chair & $\OURSst{4}$ & 7.95 & \textbf{35.67} & \textbf{1.13} & 1.62\\
            & $\OURS{8}$ & 7.70 & 37.76 & 1.87 & 1.53\\
            & $\OURSst{8}$ & 7.72 & 37.70 & 1.14 & 1.71\\
      \hline
            & $\OURS{4}$ & 6.74 & 27.22 & 0.990 & 1.69\\
      couch & $\OURSst{4}$ & 6.50 & 25.73 & \textbf{0.831} & 1.76\\
            & $\OURS{8}$ & \textbf{6.48} & 28.13 & 0.982 & 1.76\\
            & $\OURSst{8}$ & 7.32 & \textbf{24.81} & 1.12 & \textbf{1.43}\\
      \hline
            & $\OURS{4}$ & \textbf{2.28} & 22.86 & 0.919 & \textbf{2.61}\\
      plane & $\OURSst{4}$ & 2.34 & \textbf{22.40} & \textbf{0.522} & 2.87\\
            & $\OURS{8}$ & 2.32 & 23.26 & 0.886 & 2.88\\
            & $\OURSst{8}$ & 2.39 & 22.69 & 0.536 & 2.96\\
      \hline
            & $\OURS{4}$ & 4.89 & 24.06 & 0.817 & 1.59\\
      car   & $\OURSst{4}$ & 5.41 & 23.09 & 0.908 & \textbf{1.45}\\
            & $\OURS{8}$ & \textbf{4.85} & 24.29 & 0.824 & 1.60\\
            & $\OURSst{8}$ & 5.00 & \textbf{22.98} & \textbf{0.619} & 1.80\\
      \hline
    \end{tabular}
  \end{center}
\end{table}

\subsection{Qualitative Results} \label{sec:qualitative_svr}
\subsubsection{SVR on ShapeNet}
We demonstrate the qualitative results of the \textbf{SVR} task on the five ShapeNet object categories selected for our experiments in Fig.\ref{fig:samples}. As discussed in Section 5.4. of the paper, the qualitative results of the \SVR{} task are much in line with those of the \PCAE{} task. That is, the $\OURS{n}$ model brings minor stitching improvements and closes smaller holes while $\OURSst{n}$ further dramatically improves the visual quality. The reconstruction of the chair by $\OURSst{n}$ still contains undesirable holes which we found to be caused by the fact that the training in this case failed to converge within the fixed budget of training iterations which is also reflected by the CD metric in the Table \ref{tab:gridSearch_SVR} and could be in practice fixed by simply training the network for longer time.

\subsubsection{Poisson Surface Reconstruction } \label{ssec:PSR_explanation}
\begin{figure}[h]
\begin{center}
\includegraphics[width=0.8\linewidth]{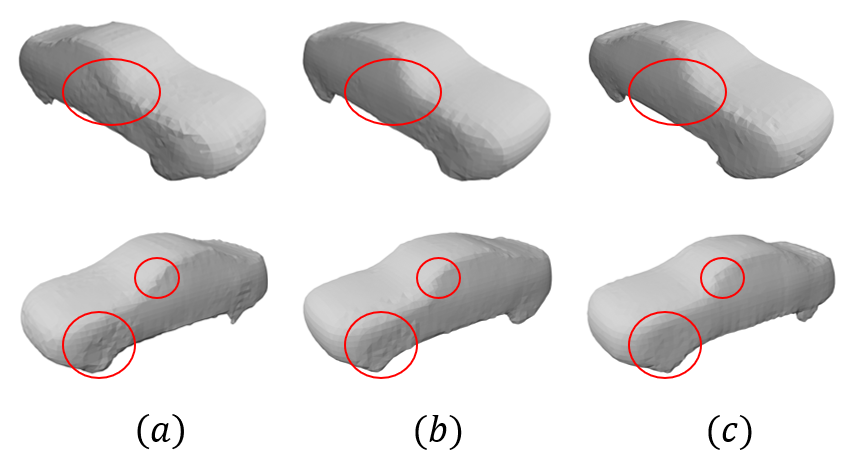}
\end{center} 
  \caption{PSR on the point cloud estimated by (a) DSP, (b) $\OURS{4}$ and (c) $\OURSst{4}$.}
\label{fig:PSR_sample}
\end{figure}

The goal of forcing consistency and reducing the gaps between the overlapping mesh patches could also be achieved by applying the Poisson Mesh Reconstruction (PSR) on the oriented vertices of the predicted points of all the mesh patches. We experimented with PSR and found that it produces less noisy meshes when applied on our outputs than on those of DSP~\cite{Bednarik20} as shown in Fig.\ref{fig:PSR_sample}.

\begin{figure*}[htb]
\centering
\centerline{\hspace{+2.5em}\includegraphics[width=0.9\textwidth]{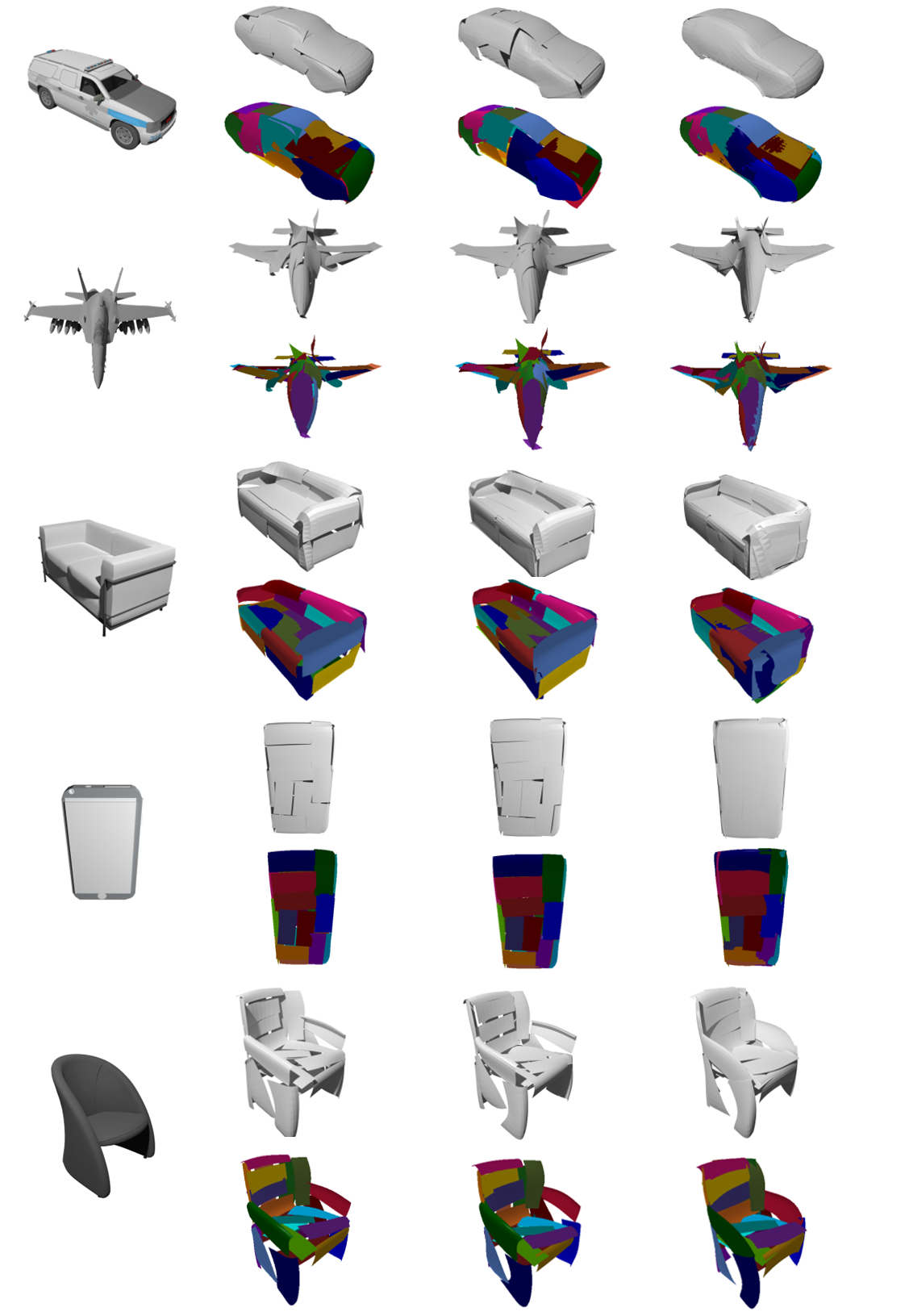}}
\caption{\textbf{Qualitative results of the \textbf{SVR} task.} From left to right: ground truth mesh (top) and the input image (bottom), DSP, $\OURS{n}$ and $\OURSst{n}$. From top to bottom: car, plane, couch, cellphone, chair.  The choice of $n$ for each category corresponds to that of Table 3 of the paper. The colors represent individual patches. The color bleeding visual artifact visible e.g. between the violet and orange patch on the nose of the right-most plane indicates that the patches nearly coincide which a sign of good stitching.}
\label{fig:samples}
\end{figure*}

\end{document}